\documentclass[sigconf,natbib=false,nonacm]{acmart}

\AtBeginDocument{%
  }

\usepackage[normalem]{ulem}
\usepackage{sidecap} 
\usepackage{amsmath,amsfonts}
\usepackage{algpseudocode}

\usepackage[ruled, vlined, norelsize, linesnumbered]{algorithm2e}
\usepackage{algorithmicx}
\usepackage{graphicx}
\usepackage{textcomp}
\usepackage{mathtools}
\usepackage{xcolor}
\usepackage{booktabs}
\SetKwInput{KwInput}{Input}
\SetKwInput{KwOutput}{Output}

\usepackage{acronym}
\usepackage{enumitem}

\acrodef{LLM}{large language model}
\acrodef{SLM}{small language model}
\acrodef{RTL}{register transfer level}
\acrodef{CVDP}{comprehensive verilog design problem}
\acrodef{SOTA}{state-of-the-art}
\usepackage{pifont}
\usepackage{soul}

\usepackage{colortbl}
\usepackage{makecell}

\definecolor{H0}{RGB}{255,235,238} % low = red
\definecolor{H1}{RGB}{255,249,196} % mid = yellow
\definecolor{H2}{RGB}{200,230,201} % high = green

\newcommand{\heat}[1]{%
    \ifdim #1pt < 10pt \cellcolor{H0}{#1}%
    \else\ifdim #1pt < 25pt \cellcolor{H1}{#1}%
    \else \cellcolor{H2}{#1}%
    \fi\fi}

\newcolumntype{G}{>{\centering\arraybackslash}c}

\usepackage{multirow}
\usepackage[caption=false,font=footnotesize,labelformat=simple]{subfig} % caption=false is needed for IEEE template
\usepackage{todonotes}
\usepackage{acronym}
\usepackage{tcolorbox}
\usepackage{balance}

\RequirePackage[
  datamodel=acmdatamodel,
  style=acmnumeric,
  ]{biblatex}

\begin{document}

%%
%% The "title" command has an optional parameter,
%% allowing the author to define a "short title" to be used in page headers.
\title{David vs. Goliath: Can Small Models Win Big with Agentic AI in Hardware Design?}

% UNCOMMENT POST PAPER RELEASE
\author[Shankar, Pandey, et al.]{%
Shashwat Shankar$^1$\textsuperscript{*}, Subhranshu Pandey$^1$\textsuperscript{*},  
Innocent Dengkhw Mochahari$^1$, Bhabesh Mali$^1$, \\
Animesh Basak Chowdhury$^2$, Sukanta Bhattacharjee$^1$, Chandan Karfa$^1$}
\affiliation{\institution{$^1$Indian Institute of Technology, Guwahati, India $^2$NXP USA, Inc.}
\country{}
}

% %%
% %% By default, the full list of authors will be used in the page
% %% headers. Often, this list is too long, and will overlap
% %% other information printed in the page headers. This command allows
% %% the author to define a more concise list
% %% of authors' names for this purpose.
% \renewcommand{\shortauthors}{Trovato et al.}

%%
%% The abstract is a short summary of the work to be presented in the
%% article.
\begin{abstract}
\Ac{LLM} inference demands massive compute and energy, making domain-specific tasks expensive and unsustainable. As foundation models keep scaling, we ask: Is bigger always better for hardware design? Our work tests this by evaluating \Acp{SLM} coupled with a curated agentic AI framework on NVIDIA's \Ac{CVDP} benchmark. Results show that agentic workflows: through task decomposition, iterative feedback, and correction - not only unlock near-\ac{LLM} performance  at a fraction of the cost but also create learning opportunities for agents, paving the way for efficient, adaptive solutions in complex design tasks.
\end{abstract}

%%
%% The code below is generated by the tool at http://dl.acm.org/ccs.cfm.
%% Please copy and paste the code instead of the example below.
%%

% \ccsdesc[500]{Do Not Use This Code~Generate the Correct Terms for Your Paper}
% \ccsdesc[300]{Do Not Use This Code~Generate the Correct Terms for Your Paper}
% \ccsdesc{Do Not Use This Code~Generate the Correct Terms for Your Paper}
% \ccsdesc[100]{Do Not Use This Code~Generate the Correct Terms for Your Paper}

%%
%% Keywords. The author(s) should pick words that accurately describe
%% the work being presented. Separate the keywords with commas.

% UNCOMMENT: PAPER ACCEPTANCE
% \setcopyright{acmlicensed}
% %        \setcopyright{none} % No copyright notice required for submissions
% \copyrightyear{2024}
% \acmYear{2024}
% \acmDOI{XXXXXXX.XXXXXXX}

% \settopmatter{printacmref=false, printccs=e, printfolios=true}
\settopmatter{printacmref=false, printccs=false, printfolios=false}
% \received{20 February 2007}
% \received[revised]{12 March 2009}
% \received[accepted]{5 June 2009}

%%
%% This command processes the author and affiliation and title
%% information and builds the first part of the formatted document.
\maketitle

%UNCOMMENT: Paper release
{
\renewcommand{\thefootnote}{\fnsymbol{footnote}}
\footnotetext[1]{S. Shankar and S. Pandey contributed equally to this work. A.B. Chowdhury, S. Bhattacharjee and C. Karfa advised equally.
}}

%UNCOMMENT: Paper release
{
\renewcommand{\thefootnote}{\fnsymbol{footnote}}
\footnotetext[2]{For a workload of 10k input tokens and 1.5k output tokens, GPT-4o-mini consumes $7.24 \pm 1.67$ Wh, whereas DeepSeek-7B only requires $0.55 \pm 0.06$ Wh~\cite{jegham2025hungryaibenchmarkingenergy}
} \label{footnote_2}}

% \textcolor{red}{JK: I've various thoughts for WIP and follow-up edits. I'll work those in locally, so please don't revise too much for now as that would trigger conflicts. Thanks!}

% !TEX root = ../main.tex
\section{Introduction}
\label{sec:introduction}

The semiconductor industry faces a critical sustainability challenge with AI-assisted design flow. Today, the industry is increasingly turning to AI for productivity gains. However, \ac{LLM}-based design automation comes at a substantial cost: GPT-4 consumes $\$0.0012$--$\$0.0016$ kWh per 1K tokens~\cite{patterson2022carbon}, translating to millions in operational expenses for large projects. At $\$0.03$ per $1K$ output tokens, processing a medium-sized SoC through iterative refinement costs $\$15K$ -- $\$45K$ for inference alone. The semiconductor industry already accounts for 2--3\% of global electricity~\cite{semiconductor_energy2023}; layering power-hungry \acp{LLM} onto design workflows exacerbates this burden precisely when sustainability pressures mount. As models scale toward trillion-parameter regimes, these costs will only intensify. This paper explores whether local, efficient \acp{SLM} viably replace or complement centralized \acp{LLM}, thus effectively redistributing demand and reducing the industry’s energy and cost footprint. Thus, the question emerges: \textit{do we always need Goliath-sized models, or can David win with the right strategy?}

\begin{figure*}[t]
\centering
\begin{minipage}{0.8\linewidth}
    \includegraphics[width=\linewidth]{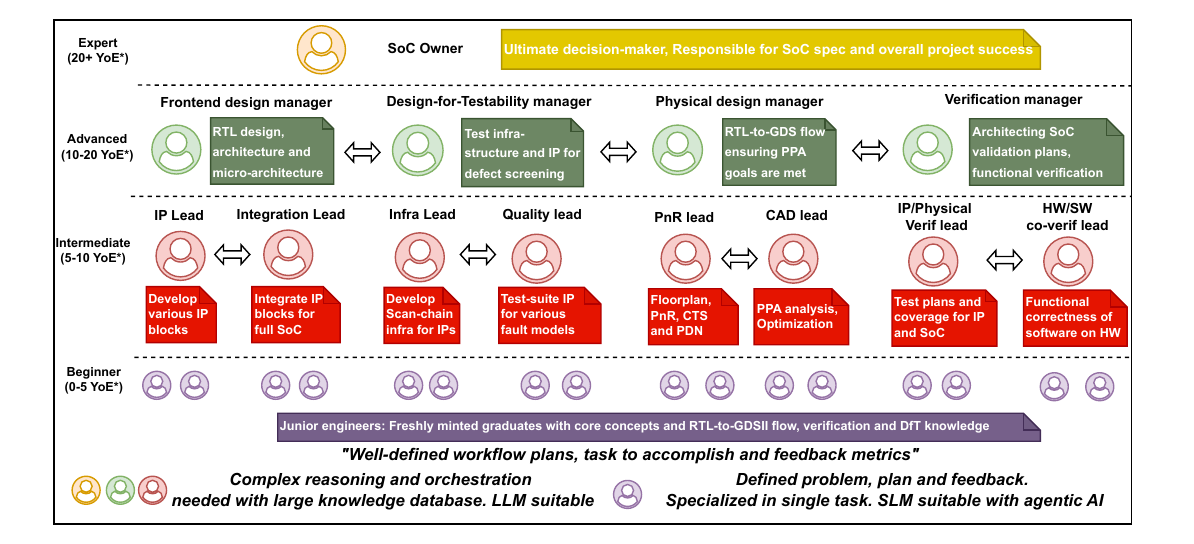}
    
\end{minipage}\hfill
\begin{minipage}{0.18\linewidth}
    \caption{Chip-design workflow for a commercial-grade SoC in a fabless semiconductor organization. YoE indicates years of experience. Task-specific SLMs integrated into a well-architected agentic-AI framework are appropriate options at beginner-level tasks to excel automation with explicit objectives, workflows, and evaluation metrics.}
    \label{fig:problem_statment}
\end{minipage}
\end{figure*}

\textbf{Organizational Analogies for AI-Assisted Design:} \autoref{fig:problem_statment} outlines how typically a fabless semiconductor chip design industry operates. Experienced engineers ($>5$ years of experience) focus on architectural decisions, technology selection, and complex timing closure -- tasks that require deep expertise. Junior engineers handle well-defined implementations: instantiating modules, connecting busses, implementing straightforward state machines, and debugging syntax errors. They succeed through clear specifications, relevant examples, immediate feedback, and quality oversight. Not every task demands senior expertise; a junior engineer perfectly implements a FIFO given clear specifications, even if unable to architect a cache coherence protocol. This organizational wisdom is lost in current automation based on \ac{LLM}. Existing systems uniformly deploy expensive large models even for routine tasks manageable by smaller alternatives, analogous to assigning principal engineers to debug syntax errors. Just as organizations cannot staff entirely with principals, sustainable AI-assisted design cannot rely exclusively on frontier \acp{LLM}.

\textbf{Research Gap:} Despite the clear need for efficient alternatives, no previous work systematically evaluates \acp{SLM} (<20B parameters) within sophisticated agentic frameworks for hardware design. Research falls into two camps: (1) Single-shot \ac{SLM} evaluation showing 35--45\% pass rates versus GPT-4's 58\%, concluding smaller models inadequate~\cite{liu2023verilogeval}, and (2) Sophisticated multi-agent systems demonstrating framework benefits but exclusively employing large commercial models~\cite{spec2rtl2025, uvllm2024}. This gap prevents understanding whether observed \ac{SLM} performance ceilings reflect inherent limitations or simply inadequate scaffolding.

\textbf{Core Hypothesis:} We hypothesize that hardware design tasks, when appropriately decomposed through agentic frameworks, create subtasks whose complexity falls within \ac{SLM} capability thresholds. Just as junior engineers succeed with proper mentorship, \acp{SLM} can achieve near-\ac{LLM} performance when equipped with sophisticated task decomposition, iterative refinement, and targeted validation. Three mechanisms enable \ac{SLM} success: (1) task decomposition breaking complex designs into subtasks, (2) structured guidance providing explicit instructions and examples, and (3) iterative refinement enabling convergence despite initial imperfections.

%Through extensive evaluation on NVIDIA's CVDP benchmark~\cite{cvdp2024}, we establish two key findings that validate this hypothesis: First, on certain hardware design tasks, agentic frameworks improve \ac{SLM} success rates by 62--141\%. DeepSeek-R1 (7B) jumps from 21\% to 51\% pass rate--- $2x$ improvement. This is because proper task decomposition makes hardware design subtasks \ac{SLM}-accessible. Second, DeepSeek-R1 (7B) achieves 57\% of GPT-o4-mini's performance at 10\% of cost, providing $10$x energy advantage (For a workload of 10k input tokens and 1.5k output tokens, GPT-4o-mini consumes $7.24 \pm 1.67$ Wh, whereas DeepSeek-7B only requires $0.55 \pm 0.06$ Wh~\cite{jegham2025hungryaibenchmarkingenergy}).

We derive two key observations from our evaluation. First, agentic task decomposition yields substantial gains for \acp{SLM}, producing 30–140\% relative improvement across code-generation tasks. The largest boost occurs in cid007, where models such as DeepSeek-R1 and Granite-4 more than double their single-shot success rates. Second, SLMs offer favorable accuracy–efficiency trade-offs in code-comprehension tasks, with models like phi-3.5-mini-instruct and DeepSeek-R1 achieving performance comparable to larger LLMs while requiring significantly fewer computational resources\ref{footnote_2}. 

Our key contributions of the paper are outlined below: 
\begin{itemize}[leftmargin=*]
\item \textbf{Comprehensive \ac{SLM}-agentic study in hardware design}: We evaluate \ac{SLM} models (1.7B--20B) across code generation and comprehension, establishing baseline performance.
\item \textbf{Novel heterogeneous agentic framework optimized for \ac{SLM} capabilities}: We crafted SLM-aware agentic AI system featuring \ac{SLM}-tailored prompt engineering, port usage validation catching incomplete implementations, error categorization for focused refinement, and rollback preventing quality regression.
\item \textbf{"Strategy over Scale"~\cite{belcak2025smalllanguagemodelsfuture} in hardware automation}: Empirical validation shows well-designed frameworks with \ac{SLM} can match and even outperform \ac{LLM} on certain hardware design tasks. 
\item \textbf{Open-source framework}: We plan to open-source our findings and analysis enabling community research on efficient AI-assisted hardware design.
\end{itemize}

The remainder is organized as: Sec.~\ref{sec:background} reviews related work and identifies the research gap; Sec.~\ref{sec:methodology} details our framework architecture; Sec.~\ref{sec:experiments} presents experimental results; Sec.~\ref{sec:conclusion}  concludes.

% \textbf{Contributions:} Our key contributions of the paper are outlined below:
% \begin{itemize}[leftmargin=*]
% \item First comprehensive \ac{SLM}-agentic study in hardware design, evaluating six models (1.7B--20B) across code generation and comprehension, establishing baseline performance characteristics.
% \item Novel heterogeneous agentic framework optimized for \ac{SLM} capabilities, featuring \ac{SLM}-tailored prompt engineering, port usage validation catching incomplete implementations, error categorization for focused refinement, and rollback preventing quality regression.
% \item Empirical validation showing well-designed frameworks with 7B models matching poorly-designed frameworks with $1T$+ models, validating "strategy over scale" in hardware automation.
% \item First systematic energy and cost comparison across model sizes, enabling informed deployment decisions: DeepSeek-R1 provides 10× cost advantage at acceptable performance.
% \item Open-source framework and analysis enabling community research on efficient AI-assisted hardware design.
% \end{itemize}

% The remainder is organized as: Section~\ref{sec:background} reviews related work and identifies the research gap; Section~\ref{sec:methodology} details our framework architecture; Section~\ref{sec:results} presents experimental results; Section~\ref{sec:discussion} discusses implications and concludes.
% !TEX root = ../main.tex
\section{Background and Related Work}
\label{sec:background}

\subsection{LLM-Based Hardware Design Automation}

The application of \acp{LLM} to hardware design has evolved through three generations. Early work focused on single-shot RTL generation from natural language. VerilogEval~\cite{liu2023verilogeval} demonstrated GPT-4 achieves 58.9\% pass@1 on basic tasks but struggles with complex, multi-module designs requiring architectural planning. CodeV~\cite{zhao2025codev}, BetterV~\cite{betterv2024} and VeriGen~\cite{thakur2023verigen} improved generation through enhanced prompting, reasoning and RAG, reaching 61.3\% accuracy.

Domain-adapted models~\cite{xu2025revolution} emerged as the second generation. ChipNeMo~\cite{chipnemo2023} demonstrated 72.3\% pass@1 on internal benchmarks through continued pre-training on proprietary hardware data, while RTLCoder~\cite{rtlcoder2024} achieved 82.3\% on RTLLM benchmark through open-source fine-tuning. \cite{li2025specllm,meng2023unlocking} and ~\cite{pulavarthi2025assertionbenchbenchmarkevaluatelargelanguage} are some of the works in LLM-aided verification domain. However, these models remain large with expensive training and deployment costs.

Current state-of-the-art employs multi-agent systems: Spec2RTL-Agent~\cite{spec2rtl2025} reduces human intervention by 75\% through iterative understanding, progressive coding, and adaptive reflection, but exclusively uses GPT-4 at substantial cost (\$300+ per implementation). UVLLM~\cite{uvllm2024} achieves 86.99\% syntax error fix rate through UVM integration. VerilogCoder~\cite{ho2024verilogcoder}, GPT4AIGChip~\cite{fu2023gpt4aigchip}, Chipmind~\cite{firouzi2025chipmnd}, ChipAlign~\cite{deng2025chipaligninstructionalignmentlarge} and ChatCPU~\cite{wang2024chatcpu} demonstrate human-LLM collaboration benefits. \cite{wang2025mcp4eda,ghose2025orfsagenttoolusingagentschip,orfs_agent} provided agentic AI framework for end-to-end RTL-to-GDSII flow. \cite{xu2025large,pan2025survey} summarize latest works in \ac{LLM}-aided design automation. 

%\textbf{Critical Gap:} All existing multi-agent systems exclusively employ large commercial models without exploring whether sophisticated agentic AI framework design targeted for smaller, efficient models to achieve comparable results.

\subsection{Small Language Models and Agentic AI}

Recent work~\cite{saadfalcon2025intelligence} introduces the metric of \textbf{"intelligence per watt"}: task accuracy per unit of power consumption—as a unified measure of both capability and efficiency for local AI inference, demonstrating that SLMs on local accelerators can now handle the majority of real-world queries at a fraction of the energy and cost of frontier models. \acp{SLM} ($\leq$20B parameters) offer compelling advantages: 8--10× lower energy consumption~\cite{jegham2025hungryaibenchmarkingenergy,touvron2023llama}, faster inference, reduced memory footprint, and lower operational costs. Recent models demonstrate impressive capabilities: Phi-3.5 (3.8B)~\cite{phi3} achieves competitive reasoning through quality training data; DeepSeek-R1 (7B)~\cite{deepseek2024} incorporates reinforcement learning for code generation; Qwen-Coder~\cite{qwen2024} specializes in programming languages including Verilog; SmolLM2 (1.7B)~\cite{smollm2024} targets edge deployment. However, \acp{SLM} face limitations: reduced reasoning capacity for multi-hop tasks, shorter context windows, and lower out-of-distribution performance~\cite{kaplan2020scaling}. Granite4~\cite{granite4_2025} is IBM’s next-generation \ac{SLM} featuring a hybrid Mamba/Transformer architecture that reduces memory and compute costs.

Agentic AI systems decompose complex tasks into manageable subtasks through iterative refinement and tool use~\cite{autogen2023, agentcoder2023}. Effective frameworks incorporate task decomposition, feedback incorporation, external tool integration, multi-agent collaboration, and rollback mechanisms~\cite{autoagents2023}. AgentCoder~\cite{agentcoder2023} demonstrates 7--15\% improvements in software through multi-agent approaches. AutoSafeCoder~\cite{autosafecoder2024} integrates static analysis and fuzz testing. However, no work systematically evaluates whether agentic scaffolding can elevate \ac{SLM} performance for hardware design.

\subsection{CVDP hardware design tasks}

The Comprehensive Verilog Design Problems (CVDP) benchmark~\cite{cvdp2024}, developed by NVIDIA, provides rigorous evaluation with 336 problems across arithmetic operations, control logic, memory systems, and miscellaneous designs. Each includes natural language specification, module interface, functional requirements, and CocoTB-based test suites. Derived from production IP blocks, it represents realistic complexity. State-of-the-art achieves only 26.5\% pass rate (GPT-4o-mini, single-shot), highlighting substantial improvement opportunity. Sec.~\ref{sec:experiments} provides detailed breakdown on the benchmark.

% \subsection{Research Positioning}

% Our work fills a critical void: systematic evaluation of \acp{SLM} within sophisticated agentic frameworks for hardware design. Unlike prior work treating model selection as binary (\ac{LLM} or nothing), we investigate the complete cost-performance spectrum. We demonstrate that thoughtful framework design enables practical tradeoffs, with 7B models achieving 57\% of GPT-4o-mini performance at 10\% of cost---transforming the sustainability calculus for AI-assisted hardware development.
% !TEX root = ../main.tex
\section{Problem Formulation}
\label{sec:problem}

We formalize our investigation into the effectiveness of \acp{SLM} for hardware design when paired with sophisticated agentic-AI frameworks. Our central question is:
\textit{Can a strategically designed, SLM-aware agentic-AI framework enable resource-efficient models to achieve practical performance in hardware design automation?}
To this end, we evaluate two core research questions through an agentic-AI based benchmarking study:

\textbf{RQ1: Can agentic frameworks elevate SLM performance to practical levels?}
Using our comprehensive evaluation pipeline on state-of-the-art CVDP benchmarks~\cite{cvdp2024}, we examine whether advanced, SLM-aware scaffolding can approach or surpass the single-shot performance of larger \acp{LLM}. We define “practical levels” as achieving functional correctness and performance sufficient for real-world deployment at attractive cost–efficiency tradeoffs.

%%%% THIS WILL BE NEEDED FOR ABLATION STUDIES
% \textbf{RQ2: Which framework components most benefit smaller models?}
% We analyze how specific agentic components—such as context-aware prompt optimization, iterative refinement, error categorization, port-level validation, and rollback mechanisms disproportionately benefit \ac{SLM} for a given task.

\textbf{2: For which specific tasks does agentic scaffolding enable SLMs to reach LLM-level performance?}
We investigate which categories of hardware-design tasks allow an \ac{SLM} equipped with sophisticated scaffolding to match or exceed \ac{LLM} performance, and which categories remain challenging where \acp{SLM} continue to underperform even with agentic support.

\section{SLM-aware agentic AI framework}
\label{sec:methodology}

% Our agentic framework explicitly accommodates \ac{SLM} constraints through external scaffolding that provides what junior engineers receive from experienced mentors: clear instructions, relevant examples, immediate feedback, and quality oversight. Figure~\ref{fig:framework} illustrates the architecture.

This section presents a structured agentic framework designed to compensate for the inherent limitations of small language models (\acp{SLM}) in hardware design automation. The framework provides the scaffolding that junior hardware engineers typically receive from senior mentors: context preparation, structured instructions, curated examples, iterative validation, and targeted feedback.
Figure~\ref{fig:framework} illustrates the system architecture, consisting of five cooperating agents that transform design intent from the CVDP dataset into verified \ac{RTL} implementations.

% \begin{figure*}[t]
%     \centering
%     \includegraphics[width=0.7\textwidth]{figures/david_vs_goliath_flow_diagram_v1.pdf}
%     \caption{Proposed SLM-aware agentic AI framework. (1) Planning and Pre-processing Agent retrieves and structures context from the CVDP dataset; (2) SLM-Optimized Prompt Engineering Agent constructs SLM-aware prompts using keyword injection, in-context examples, and token budgeting; (3) CodeGen Agent generates candidate RTL implementations; (4) Validation Agent performs syntax checking, I/O port usage analysis, and functional testing; and (5) Adaptive Feedback Agent categorizes errors, evaluates quality, and produces structured refinement prompts. These agents form a closed-loop iterative workflow.}
%     \label{fig:framework}
% \end{figure*}

\begin{figure*}[t]
\centering
\begin{minipage}{0.55\linewidth}
    \includegraphics[width=\linewidth]{figures/david_vs_goliath_flow_diagram_v1.pdf}
\end{minipage}\hfill
\begin{minipage}{0.42\linewidth}
    \caption{Proposed SLM-aware agentic AI framework. (1) PPA retrieves and structures context from the dataset; (2) SPEA constructs SLM-aware prompts using keyword injection, in-context examples, and token budgeting; (3) CA generates candidate RTL implementations; (4) VA performs syntax checking, I/O port usage analysis, and functional testing; and (5) AFA categorizes errors, evaluates quality, and produces structured refinement prompts. These agents form a closed-loop iterative workflow.}
    \label{fig:framework}
\end{minipage}

\end{figure*}

\subsection{Planning and Pre-processing Agent (\texttt{PPA})}

The \textit{Planning and Pre-processing Agent} is the pipeline entry point (leftmost block in Figure~\ref{fig:framework}) and prepares task-relevant context and metadata for effective \ac{SLM} operation. It has three coordinated components:
\begin{itemize}[leftmargin=*]
    \item \textbf{Context Retriever}: Extracts relevant design descriptions, interface specifications, and auxiliary documentation from the CVDP dataset using keyword- and structure-based retrieval.
    \item \textbf{Planner}: Decomposes the design problem into sub-goals behavioral intent, interface semantics, and corner-case conditions—to provide structured guidance to downstream agents.
    \item \textbf{Database Generator}: Packages task descriptions, retrieved context, and planning metadata into a unified context bundle.
\end{itemize}

This agent grounds downstream prompts in complete, accurate context while keeping \acp{SLM} lightweight.

\subsection{SLM-aware Prompt Engineering Agent (\texttt{SPEA})}

The \textit{SLM-aware Prompt Engineering Agent}, represented in Figure~\ref{fig:framework}, constructs prompts that maximize \ac{SLM} reliability through strategic structuring and explicit instructional cues. It addresses \ac{SLM} challenges in instruction following, long-context reasoning, and token efficiency. The agent integrates three submodules:
\begin{itemize}[leftmargin=*]
    \item \textbf{SLM-Aware Keyword Injector}:
Injects deterministic structural keywords—
\texttt{ROLE} (establishes expert identity), \texttt{TASK} (delineates objective), \texttt{REQUIREMENTS} (mandatory constraints), \texttt{CONSTRAINTS} (technical limitations), and \texttt{GENERATE} (requests code output). We introduced these optimized prompts motivated by \cite{lester2021power,li2021prefix} to optimize tokens consumed. These serve as attention anchors for \acp{SLM}, enhancing hierarchical instruction parsing.
  \item \textbf{In-Context Learner}: Selects few-shot examples (e.g., counter, FIFO, FSM) matched by task similarity. Examples are incorporated into the prompt to serve as pattern templates analogous to senior engineer design references.
  \item \textbf{Token Budget Manager}: Allocates limited context capacity (typically $8$K–$32$K tokens) across task descriptions ($40\%$), context files ($40\%$), and examples ($20\%$). Large files are truncated using structure-preserving heuristics to retain essential information.
\end{itemize}

We carefully engineered this agent to generate a fully structured, SLM-optimized prompt that guides subsequent code generation with minimal ambiguity and fairly deterministic output.

\subsection{CodeGen Agent (\texttt{CA})}

The \textit{CodeGen Agent} performs reasoning-guided Verilog generation using the structured prompt produced by the SLM-aware prompt engineering agent. Its responsibilities include: 1) Executing deterministic \ac{SLM}-based code generation. 2) 
Suppressing verbose natural-language reasoning to maximize available output tokens for code generation. 3) Extracting candidate Verilog modules using pattern-matched parsing. and 4) Preparing the candidate implementation for multi-stage validation. This agent embodies the core generative function of the framework while relying on external scaffolding to steer \ac{SLM} behavior.

% \subsection{Validation Agent (\texttt{VA})}

% The \textit{Validation Agent}, enforces correctness through a three-stage validation pipeline. Its primary objective is to provide the incremental validation that senior engineers typically conduct during code review.:

% \begin{itemize}[leftmargin=*]
%     \item Syntax Validation: This detects parse errors, type mismatches, undeclared identifiers, and structural inconsistencies by using a typical RTL code linter.
%     \item I/O Port Usage Checker: A common \ac{SLM} failure mode generates syntactically correct modules that pass compilation but omit functionality. Often declared I/O ports are never used in logic. These "zombie ports" indicate incomplete implementations. Our port usage analyzer performs three-phase validation: (1) extract all input/output port declarations using regex patterns handling various styles (ANSI, non-ANSI, SystemVerilog, Verilog-2001), (2) detect usage by scanning module body for port references in expressions (inputs) or assignments (outputs via non-blocking, blocking, or continuous assignment), and (3) generate targeted feedback for unused/unassigned ports.
%     \item CocoTB Test Runner: Conducts functional testing across nominal and corner-case scenarios. Test success terminates the pipeline; test failure invokes corrective feedback via the Adaptive Feedback Agent.
% \end{itemize}

\subsection{Validation Agent (\texttt{VA})}

The \textit{Validation Agent} enforces correctness through a three-stage pipeline, emulating incremental checks typically performed by senior engineers during code review:

\begin{itemize}[leftmargin=*]
    \item \textbf{Syntax Validation}: Detects parse errors, type mismatches, undeclared identifiers, and structural inconsistencies using an RTL code linter.
    \item \textbf{I/O Port Usage Checker}: Identifies “zombie ports”—declared but unused I/Os—by (1) extracting all port declarations via regex (ANSI, non-ANSI, SystemVerilog, Verilog-2001), (2) scanning the module body for input/output usage, and (3) generating targeted feedback for unused or unassigned ports.
    \item \textbf{CocoTB Test Runner}: Executes functional tests across nominal and corner-case scenarios. Success terminates the pipeline; failure triggers corrective feedback via Adaptive Feedback Agent.
\end{itemize}

\subsection{Adaptive Feedback Agent (\texttt{AFA})}

The \textit{Adaptive Feedback Agent} generates structured refinement signals and enables error-aware iteration. Its components include:

\begin{itemize}[leftmargin=*]
    \item \textbf{Error Categorizer}: Maps simulator and validator messages to one of seven categories:
syntax, undeclared, type, width, latch, timing, or novel. This structured categorization ensures targeted refinement instructions for \ac{SLM}.
    \item \textbf{Quality Scorer}: Produces a composite score combining functional correctness ($60\%$), compilation quality ($20\%$), port completeness ($10\%$), and code structure ($10\%$). Lower-scoring updates trigger rollback to the highest-scoring version.
    \item \textbf{Contextual Error Gatherer}: Synthesizes concise error summaries (e.g., line-level syntax issues, missing assignments) and integrates them into refinement prompts, including “damage repair” history to prevent repeated failed attempts.
\end{itemize}
This agent enables targeted, context-dependent correction, analogous to senior engineer feedback during iterative design refinement.

\subsection{Putting It All Together}

Figure~\ref{fig:framework} shows the full agentic AI workflow as a closed-loop pipeline. The \texttt{PPA} agent retrieves and structures task context from the dataset and generates a plan. The \texttt{SPEA} agent builds an SLM-aware prompt with keyword injection, curated examples, and context constraints. The \texttt{CA} agent produces initial Verilog implementations. The \texttt{VA} agent evaluates candidates for syntax, port usage, and functional correctness. Syntax or port errors trigger structured error prompts; functional test failures invoke the \texttt{AFA} agent to generate category-specific corrective prompts.  

This pipeline iterates prompt engineering, code generation, and validation for up to five rounds, guided by scoring, rollback, and error-awareness, until all checks pass, the iteration limit is reached, or early termination criteria are met. This coordinated loop enables \acp{SLM} to produce high-quality, functionally correct \ac{RTL} despite limited internal reasoning capabilities.

% !TEX root = ../main.tex
\section{Experimental Evaluation}
\label{sec:experiments}

\subsection{Experimental Setup}
We now describe the benchmark suite, task selection criteria, and model configurations used in our evaluation.

\subsubsection{Benchmark Suite}
Our experiments are conducted on the \emph{Comprehensive Verilog Design Problems (CVDP)} benchmark suite~\cite{cvdp2024}, a recently released and systematically curated dataset targeting RTL design, debugging, and verification tasks. CVDP organizes problems into three high-level categories:

\begin{itemize}[leftmargin=*]
    \item \textbf{Non-Agentic Code Generation:} Includes RTL module creation, logic completion, micro-architectural construction, and code repair. Subtasks cover combinational/sequential design, FSM generation, structural code fixing, signal-width correction, latch removal, and general debugging.
    
    \item \textbf{Non-Agentic Code Comprehension:} Focuses on extracting functional intent from HDL artifacts. Representative subtasks include RTL/Testbench correspondence, identifying behavioral mismatches, natural-language question answering about design intent, and summarization of module functionality.
    
    \item \textbf{Agentic Code Generation:} Involves multi-step tasks requiring planning, iterative correction, and tool-driven refinement such as synthesis-guided debugging and simulation-based repair.
\end{itemize}

CVDP is specifically designed to probe reasoning, syntax discipline, hardware-awareness, and tool-aligned correctness. As reported in~\cite{cvdp2024}, state-of-the-art LLMs achieve only \emph{34\% pass@1} on Non-Agentic Code Generation and perform significantly worse on several Code Comprehension subtasks, highlighting the inherent difficulty of the benchmark.

For a controlled and meaningful evaluation of our agentic framework, we limit our experiments to the \emph{Non-Agentic Code Comprehension} and \emph{Non-Agentic Code Generation} categories (\autoref{tab:cvdp_curated_dataset}). We intentionally exclude CVDP’s Agentic Code Generation tasks because even frontier LLMs were shown to underperform on these tasks in single-shot settings, making it difficult to isolate the gains attributable to our agentic scaffolding. For Code Comprehension tasks, we evaluate both \acp{LLM} and \acp{SLM} strictly in single-shot mode. For Code Generation tasks, we report: 1) Single-shot (SS) correctness (\emph{pass@1} with $n=5$ completions), and  2) Agentic (AG) performance using our full refinement loop (iteration cap = 5).

\begin{table}[t]
\centering
\footnotesize
\setlength{\tabcolsep}{4pt}

\caption{CVDP dataset considered for SLM evaluation}

\begin{tabular}{l p{4.8cm} c}
\toprule
\textbf{ID} & \textbf{Category Description} & \textbf{Non-agentic problems} \\
\midrule
\multicolumn{3}{l}{\textbf{Code Generation}} \\
\midrule

\textbf{cid02} & RTL – Code Completion & 94 \\

\textbf{cid03} & RTL – Natural Language Specification to Code & 78 \\

\textbf{cid04} & RTL – Code Modification & 56 \\

\textbf{cid07} & RTL – Code Improvement (Linting / QoR) & 41 \\

\textbf{cid16} & Design Verification – Debugging / Bug Fixing & 36 \\

\midrule
\multicolumn{3}{l}{\textbf{Code Comprehension}} \\
\midrule

\textbf{cid06} & Correspondence – RTL to/from Specification & 34 \\

\textbf{cid08} & Correspondence – Testbench to/from Test Plan & 29 \\

\textbf{cid09} & Question \& Answer – RTL & 34 \\

\textbf{cid10} & Question \& Answer – Testbench & 26 \\

\bottomrule
\end{tabular}
\label{tab:cvdp_curated_dataset}
\end{table}

\subsubsection{Models Evaluated}
We benchmark four resource-efficient \acp{SLM} widely used for code generation tasks: SmolLM2 (1.7B), Nemotron-Mini (4B), Granite-4 (3B), and DeepSeek-R1 (7B).  For code comprehension, we additionally evaluated with GPT-OSS (20B) and Phi3.5-mini-construct. For comparison against a frontier model, we include GPT-o4-mini as the \ac{LLM} baseline. All models use temperature $0.7$ for generation-oriented tasks to preserve architectural diversity and temperature $0.3$ for interface-heavy outputs to prioritize determinism and structural precision. A global timeout of $300$ seconds is enforced per test case. To measure the relative importance of framework components, we evaluate two configurations:  
\emph{(1) single-shot without scaffolding},  
\emph{(2) full agentic framework with adaptive refinement}.
Experiments were executed on an AMD EPYC 7763 (2.45\,GHz) server with 512\,GB RAM and NVIDIA L40S GPUs.

%  \subsection{Observations}

% Three key observations based on the experienced-junior engineer analogy:

% \textbf{H1: Task decomposition reduces complexity below SLM thresholds.} Breaking complex designs into focused subtasks (interface $\rightarrow$ data path $\rightarrow$ control $\rightarrow$ testbench) reduces per-subtask complexity such that 80--90\% fall within \ac{SLM} capabilities. While complete cache controllers exceed 4B model capabilities, individual components (tag comparator, LRU logic) may not.

% \textcolor{red}{\textbf{H2: Agentic frameworks benefit SLMs more than LLMs.} We hypothesize differential impact: \acp{SLM} show 50--150\% improvement (agentic vs. single-shot) while \acp{LLM} show 0--30\% or degradation. \acp{LLM} possess strong internal reasoning that external scaffolding may hinder; \acp{SLM} lack this and benefit from guidance.}

% \textbf{H3: Strategic framework design rivals raw scaling}: Well-designed frameworks with models less than $30$B can match or exceed poorly-designed frameworks with trillion parameters models---the "David vs. Goliath" paradigm validating that architecture matters more than scale.

\subsection{Results}

\subsubsection{Performance on Code Generation Tasks}

The overall performance of both the single-shot and agentic configurations on the CVDP Code Generation benchmark is summarized in ~\autoref{tab1:codegen_success} and ~\autoref{tab1:codegen_breakdown}. Table~\ref{tab1:codegen_success} reports the pass@1 metric across all evaluated SLMs and LLMs. Table~\ref{tab1:codegen_breakdown} further decomposes, for the best outcome between single-shot and agentic generation, the number of submissions that successfully pass the testbench, those exhibiting functional bugs, and cases with syntactic errors.

\textbf{Insights:} The results in \autoref{tab1:codegen_success} reveal a notable trend, particularly for problem category cid007. All evaluated SLMs demonstrate substantial gains when paired with the agentic framework, with \uline{DeepSeek-r1 and Granite-4 even surpassing the performance of GPT-o4 mini used as the agentic core} ($51.25\%$,$48.75\%$ vs. $44.74\%$). Correspondingly, the breakdown in ~\autoref{tab1:codegen_breakdown} shows that, for nearly all SLMs, the number of test-passing outputs for cid007 approximately improve by 100\%(relative to single-shot generation performance as baseline), achieving parity with \ac{LLM} performance. This indicates that specific hardware design tasks can be effectively handled by \acp{SLM} without sacrificing solution quality, offering significant opportunities for reducing energy and cost overheads.

For category cid004, we observe that most \acp{SLM} fail to produce functionally correct solutions (with the exception of Granite-4 under the agentic setup). However, Table~\ref{tab1:codegen_breakdown} provides additional granularity: it distinguishes functional failures from purely syntactic ones, revealing that single-shot models produce far more syntactic errors than their agentic counterparts. This again highlights the agentic framework’s stabilizing influence on \ac{SLM} code quality.

Apart from a small set of anomalies (e.g., Nemotron and Granite-4 on cid002, cid003, and cid016), the agentic workflow consistently improves SLM performance across categories. While exploring task-specialized \acp{SLM} combined with agentic pipelines is left for future work, the objective of this study is not to outperform LLMs. Rather, our aim is to identify hardware-design tasks where SLMs—augmented with agentic reasoning—offer a compelling efficiency–accuracy trade-off. In this context, we find that category cid007 is particularly well-suited for \ac{SLM} deployment.

\subsubsection{Performance on Code Comprehension Tasks}
~\autoref{tab3:code_comprehension} presents the performance of \acp{SLM} and \acp{LLM} on Code Comprehension task. Categories cid006 and cid008 require models to produce code or structured outputs adhering to a predefined schema; these tasks are evaluated using BLEU scores. In contrast, cid009 and cid010 are assessed using subjective correctness judgments provided by an \ac{LLM} judge (GPT-5-mini), following the same protocol used in the main CVDP evaluation.

\textbf{Insights:} The results in \autoref{tab3:code_comprehension} reveal several noteworthy patterns. First, \acp{SLM} demonstrate unexpectedly strong performance in higher-level reasoning tasks such as cid009 and cid010. \uline{Both phi-3.5-mini-instruct and deepseek-r1 achieve accuracy comparable to, and in some cases matching, LLM performance (e.g., $82$-$92\%$ on cid009/cid010)}. This suggests that specific forms of semantic understanding in hardware-related code analysis do not necessarily require large-scale models; rather, they benefit from model architectures optimized for structured reasoning.

For tasks requiring structured code reconstruction (cid006 and cid008), several \acp{SLM} outperform LLM baselines. \uline{Notably, gpt-oss achieves the highest score on cid006, and both phi-3.5-mini-instruct and deepseek-r1 substantially exceed LLM performance on cid008.} This trend indicates that the SLM's inductive bias toward code-token distributions may be advantageous when producing deterministic, schema-constrained output. 

The Testbench statistics further reinforce this observation: while overall pass rates vary across models, \ac{SLM} such as phi-3.5-mini-instruct and deepseek-r1 outperform larger models like GPT-4o in total successful cases.\ indicating code comprehension tasks are well-suited for \acp{SLM}. Fig.~\ref{fig:code_compre_response} presents one such case where the same prompt have been given to GPT-o4-mini and DeepSeek-R1 and GPT-o4-mini ended up just producing code, whereas DeepSeek-R1 produced the detailed \texttt{subjective.txt} as instructed in the prompt.

% \begin{table}[t]
% \centering
% \caption{Code generation performance: single-shot vs. agentic framework.}
% \label{tab:overall_performance}
% \small
% \begin{tabular}{l|c|r|r|r}
% \toprule
% \textbf{Model} & \textbf{Params} & \textbf{Single} & \textbf{Agentic} & \textbf{Improve} \\
% \midrule
% SmolLM2 & 1.7B & 16 (4.7\%) & 26 (7.6\%) & +62.5\% \\
% Phi-3.5 & 3.8B & 15 (4.4\%) & -- & -- \\
% Nemotron & 4B & 22 (6.4\%) & 29 (8.5\%) & +31.8\% \\
% DeepSeek & 7B & 17 (5.0\%) & 41 (12.0\%) & \textbf{+141.2\%} \\
% Phi-4 & 14B & -- & -- & -- \\
% GPT-OSS & 20B & -- & -- & -- \\
% \midrule
% GPT-5-mini & -- & 89 (26.5\%) & 72 (21.2\%) & -19.1\% \\
% \bottomrule
% \end{tabular}
% \end{table}

\begin{table}[t]

\setlength\tabcolsep{3pt}
\caption{SLMs/LLMs with single-shot(SS) and agentic framework(AG) on CVDP non-agentic code generation dataset.
Green/yellow/red denote high/medium/low performance.}

\resizebox{0.49\textwidth}{!}{
\centering
%\fontsize{6pt}{7pt}\selectfont
\scriptsize
\begin{tabular}{l c *{10}{G}}
\toprule

\multirow{2}{*}{Model} & 
\multirow{2}{*}{\begin{tabular}{c} Total\\ tests \end{tabular}} &
\multicolumn{10}{c}{Code Generation (Pass$@$1 \% with n=5)} \\
\cmidrule(lr){3-12}

 &  & \multicolumn{2}{c}{cid002} & \multicolumn{2}{c}{cid003} & \multicolumn{2}{c}{cid004} & \multicolumn{2}{c}{cid007} & \multicolumn{2}{c}{cid016} \\
\cmidrule(lr){3-4} \cmidrule(lr){5-6} \cmidrule(lr){7-8} \cmidrule(lr){9-10} \cmidrule(lr){11-12}

& &
SS & AG &
SS & AG &
SS & AG &
SS & AG &
SS & AG \\
\midrule

Nemotron-Mini & 342 &
\heat{4.26} & \heat{0} &
\heat{1.28} & \heat{0} &
\heat{0} & \heat{0} &
\heat{20} & \heat{36} &
\heat{2.86} & \heat{0} \\

SmolLM & 342 &
\heat{1.03} & \heat{1.03} &
\heat{0} & \heat{0} &
\heat{0} & \heat{0} &
\heat{18.75} & \heat{30} &
\heat{0} & \heat{0} \\

DeepSeek-R1  & 342 &
\heat{0} & \heat{0} &
\heat{0} & \heat{0} &
\heat{0} & \heat{0} &
\heat{21.25} & \heat{51.25} &
\heat{0} & \heat{0} \\

Granite-4  & 342 &
\heat{9.57} & \heat{0} &
\heat{6.41} & \heat{0} &
\heat{0} & \heat{1.82} &
\heat{20.51} & \heat{48.75} &
\heat{2.86} & \heat{0} \\

\midrule
GPT-o4 Mini & 342 &
\heat{17.02} & \heat{24.47} &
\heat{30.77} & \heat{17.95} &
\heat{20} & \heat{12.73} &
\heat{41} & \heat{44.74} &
\heat{20} & \heat{22.86} \\

\bottomrule
\end{tabular}
}
\label{tab1:codegen_success}
\end{table}

\begin{table}

\centering
\scriptsize
\setlength{\tabcolsep}{2pt}
\caption{Breakdown of Code generation problem generated by single shot and agentic framework. \texttt{P}: Testbench passing, \texttt{FE}: Functional error, \texttt{SE}: Syntactical error }
\resizebox{0.49\textwidth}{!}{
\begin{tabular}{l l *{15}{c}}
\toprule
\multirow{2}{*}{Model} & \multirow{2}{*}{Mode} &
\multicolumn{3}{c}{cid002} &
\multicolumn{3}{c}{cid003} &
\multicolumn{3}{c}{cid004} &
\multicolumn{3}{c}{cid007} &
\multicolumn{3}{c}{cid016} \\
\cmidrule(lr){3-5} \cmidrule(lr){6-8} \cmidrule(lr){9-11}
\cmidrule(lr){12-14} \cmidrule(lr){15-17}
& & P & FE & SE & P & FE & SE & P & FE & SE & P & FE & SE & P & FE & SE \\
\midrule

\multirow{2}{*}{\begin{tabular}{l} Nemotron\\ mini \end{tabular}} & SS & 4 & 22 & 68 & 1 & 16 & 61 & 0 & 15 & 40 & 16 & 30 & 34 & 1 & 7 & 27 \\
                              & AG & 0 & 10 & 84 & 0 & 7 & 71 & 0 & 38 & 17 & 29 & 48 & 4 & 0 & 35 & 0 \\
\midrule

SmolLM & SS & 1 & 14 & 79 & 0 & 10 & 68 & 0 & 13 & 42 & 15 & 33 & 32 & 0 & 12 & 23 \\
       & AG & 1 & 28 & 68 & 0 & 22 & 66 & 0 & 28 & 28 & 25 & 48 & 11 & 0 & 27 & 9 \\
\midrule

Deepseek-r1 & SS & 0 & 10 & 84 & 0 & 6 & 72 & 0 & 9 & 46 & 17 & 24 & 39 & 0 & 5 & 30 \\
            & AG & 0 & 94 & 0 & 0 & 78 & 0 & 0 & 55 & 0 & 41 & 39 & 0 & 0 & 35 & 0 \\
\midrule
Granite4 & SS & 9 & 33 & 52 & 5 & 39 & 34 & 0 & 15 & 40 & 16 & 31 & 31 & 1 & 15 & 19 \\
         & AG & 0 & 10 & 84 & 0 & 6 & 72 & 1 & 37 & 17 & 39 & 33 & 4 & 0 & 35 & 0 \\

\midrule

         \multirow{2}{*}{\begin{tabular}{l} GPT-o4\\ mini \end{tabular}} & SS & 16 & 51 & 27 & 24 & 42 & 12 & 11 & 31 & 13 & 31 & 36 & 7 & 7 & 26 & 2 \\
                           & AG & 23 & 26 & 45 & 14 & 34 & 30 & 7 & 35 & 13 & 34 & 38 & 4 & 8 & 22 & 5 \\
\bottomrule
\end{tabular}
}
\label{tab1:codegen_breakdown}
\end{table}

\begin{table}

\setlength\tabcolsep{3pt}
\caption{SLMs/LLMs performance: Code comprehension}
\resizebox{0.49\textwidth}{!}{
\centering
\scriptsize
\begin{tabular}{l c*{7}{G}}
\toprule

\multirow{2}{*}{Model} & 
\multicolumn{3}{c}{Testbench stats} &
\multicolumn{4}{c}{Code Comprehension (Pass$@$1)} \\
\cmidrule(lr){2-4} \cmidrule(lr){5-8}

 & Total & Pass & (\%) & cid006 & cid008 & cid009 & cid010 \\
\midrule

phi3.5-mini-instruct& 186 & \heat{106} & \heat{56.99} &
\heat{47.06} & \heat{37.93} & \heat{82.35} & \heat{92.31} \\

deepseek-r1 & 186 & \heat{106} & \heat{56.99} &
\heat{50.00} & \heat{37.93} & \heat{76} & \heat{92} \\

gpt-oss & 186 & \heat{94} & \heat{50.54} &
\heat{58.82} & \heat{37.93} & \heat{47} & \heat{62} \\

nemotron-mini & 186 & \heat{62} & \heat{33.33} &
\heat{29.41} & \heat{10.34} & \heat{64.71} & \heat{53.85} \\

smollm2 & 186 & \heat{14} & \heat{7.53} &
\heat{11.76} & \heat{10.34} & \heat{0} & \heat{0} \\
\midrule
GPT-4o & 186 & \heat{74} & \heat{39.78} &
\heat{33.33} & \heat{11.11} & \heat{91.18} & \heat{96.15} \\

GPT-3.5-turbo & 186 & \heat{61} & \heat{32.80} &
\heat{18.18} & \heat{0} & \heat{82.35} & \heat{96.15} \\

GPT-o4 Mini & 186 & \heat{61} & \heat{32.80} &
\heat{13.64} & \heat{6} & \heat{82.35} & \heat{96.15} \\

\bottomrule
\end{tabular}
}
\label{tab3:code_comprehension}
\end{table}

\begin{figure*}[t]
\centering
\begin{minipage}[t]{0.49\textwidth}
\scriptsize
\textbf{GPT-o4-mini Response (NO subjective.txt file generated)}
\begin{tcolorbox}[colback=white,colframe=black!40,boxrule=0.2pt]
[GPT-4o-mini has NO subjective.txt - REJECTED by scoring]
\begin{verbatim}
Output structure: ['verif/tb_qam16_mapper_interpolated.sv']

First output (verif/tb_qam16_mapper_interpolated.sv):
// Function to compute expected outputs (model of the system)
task model_system();
    logic signed [OUT_WIDTH-1:0] mapped_I [0:N-1];
    logic signed [OUT_WIDTH-1:0] mapped_Q [0:N-1];
    logic signed [OUT_WIDTH:0] interp_I [0:N/2-1];
    logic signed [OUT_WIDTH:0] interp_Q [0:N/2-1];

    // Map symbols
    for (int i = 0; i < N; i++) begin
        case (test_symbols[i][3:2])
            2'b00: mapped_I[i] = -3;
            2'b01: mapped_I[i] = -1;
            2'b10: mapped_I[i] = 1;
            2'b11: mapped_I[i] = 3;
        endcase
        case (test_symbols[i][1:0])
            2'b00: mapped_Q[i] = -3;
            2'b01: mapped_Q[i] = -1;
            2'b10: mapped_Q[i] = 1;
            2'b11: mapped_Q[i] = 3;
        endcase
    end
\end{verbatim}
\end{tcolorbox}

\scriptsize
\textbf{DeepSeek-R1 (7B) Response (Subjective.txt)}

\begin{tcolorbox}
[colback=white,colframe=black!40,boxrule=0.2pt]
[DeepSeek-R1 has subjective.txt field - ACCEPTED by scoring]

\raggedright

Okay, so I need to figure out which part of the testbench file \texttt{tb\_qam16\_mapper\_interpolated.sv} implements the Golden Model as described in the test plan. Let me start by understanding what the Golden Model is. From the test plan, the Golden Model is a reference implementation that mimics the DUT's expected behavior using math or algorithms. It maps input symbols to I and Q values and interpolates between them.

Looking at the testbench, I see there's a task called \texttt{`model\_system()`}. Let me read through what it does. It initializes some logic variables and then loops through each symbol. For each symbol, it maps the bits to I and Q values. It uses a case statement based on the bits of the symbol to assign values like -3, -1, 1, 3 for both I and Q. That seems like the mapping part.

After mapping, it does interpolation. It loops through every two symbols, takes their mapped I and Q values, averages them by shifting right by 1 (which is equivalent to dividing by 2), and stores those in \texttt{$`interp\_I`$} and \texttt{$`interp\_Q`$}. This interpolation step is crucial because the DUT probably uses some form of interpolation, so the model needs to emulate that.

Then, it builds the output by interleaving the mapped symbols with the interpolated values. For each pair of symbols, it places the first symbol, then the interpolated value, then the second symbol. This creates the output sequence that the DUT would produce.

So, the \texttt{$`model\_system()`$}task is doing all the mapping and interpolation as described in the test plan's Golden Model. It's creating the expected I and Q outputs that the DUT should match. Therefore, this task is the Golden Model used for verification.

I should also check if there's any other part that might be doing similar things, but from what I can see, \texttt{$`model\_system()`$} is the main function that implements the reference model. The \texttt{$`check\_outputs()`$} task then compares the DUT's outputs against this model's results.

\end{tcolorbox}
\end{minipage}\hfill
\begin{minipage}[t]{0.49\textwidth}
\scriptsize
\textbf{DeepSeek-R1 (7B) (contd.)}
\begin{tcolorbox}[colback=white,colframe=black!40,boxrule=0.2pt]
\begin{verbatim}

So, putting it all together, the section implementing the Golden Model is the 
\texttt{$`model\_system()`$} task in the testbench.
</think>

The section in the \texttt{$`tb\_qam16\_mapper\_interpolated`$} testbench that

implements the Golden Model is the \texttt{$`model\_system()`$} task. This task:

1. Maps each input symbol to its corresponding in-phase (I) and quadrature (Q) 
values based on the specification.
2. Interpolates between pairs of mapped symbols to emulate the DUT's interpolation 
logic.
3. Constructs the expected output sequence by interleaving mapped symbols with 
interpolated values.

Here is the relevant code:

```verilog
// Function to compute expected outputs (model of the system)
task model_system();
    logic signed [OUT_WIDTH-1:0] mapped_I [0:N-1];
    logic signed [OUT_WIDTH-1:0] mapped_Q [0:N-1];
    logic signed [OUT_WIDTH:0] interp_I [0:N/2-1];
    logic signed [OUT_WIDTH:0] interp_Q [0:N/2-1];

    // Map symbols
    for (int i = 0; i < N; i++) begin
        case (test_symbols[i][3:2])
            2'b00: mapped_I[i] = -3;
            2'b01: mapped_I[i] = -1;
            2'b10: mapped_I[i] = 1;
            2'b11: mapped_I[i] = 3;
        endcase
        case (test_symbols[i][1:0])
            2'b00: mapped_Q[i] = -3;
            2'b01: mapped_Q[i] = -1;
            2'b10: mapped_Q[i] = 1;
            2'b11: mapped_Q[i] = 3;
        endcase
    end

    // Interpolate
    for (int i = 0; i < N/2; i++) begin
        interp_I[i] = (mapped_I[2*i] + mapped_I[2*i+1]) >>> 1;
        interp_Q[i] = (mapped_Q[2*i] + mapped_Q[2*i+1]) >>> 1;
    end

    // Build output
    for (int i = 0; i < N/2; i++) begin
        model_I[i*3]     = mapped_I[2*i];
        model_I[i*3 + 1] = interp_I[i];
        model_I[i*3 + 2] = mapped_I[2*i+1];

        model_Q[i*3]     = mapped_Q[2*i];
        model_Q[i*3 + 1] = interp_Q[i];
        model_Q[i*3 + 2] = mapped_Q[2*i+1];
    end
endtask
\end{verbatim}
\end{tcolorbox}
\end{minipage}
\caption{GPT-o4 mini vs. Deepseek-r1 (7B) response for the same prompt (Problem ID: \texttt{$cvdp\_copilot\_16qam\_mapper\_0004
$).}}
\label{fig:code_compre_response}
\end{figure*}

\section{Conclusion}
\label{sec:conclusion}

% In this work, we present the first comprehensive study of \ac{SLM}-centric agentic AI system on the state-of-the-art hardware design tasks proposed in \ac{CVDP} dataset. We did this study from a sustainable AI perspective to deep-dive and understand what types of tasks can be precisely outlined, appropriately broken down and well-defined metrics can be solved using \ac{SLM} (much like task assigned to a junior level engineer). We ultimately identify the flavour of tasks which can reach near-LLM performance and if appropriate scaffolding is done can even achieve better performance. Our work paves the way that task specific SLM can later be developed and with appropriate scaffolding can reach near-LLM performance paving the way for sustainable and energy efficient AI-based hardware design.

In this work, we present the first systematic study of SLM-centric agentic AI systems applied to state-of-the-art hardware design tasks from the CVDP benchmark. Framing hardware design workflows through the lens of organizational hierarchy, we show that certain well-defined subtasks—similar to work handled by beginners—can be effectively performed by compact, energy-efficient SLMs.

Our experiments demonstrate that, with appropriate scaffolding via agentic frameworks—including task decomposition, structured guidance, and iterative refinement—SLMs can achieve near-LLM performance on both code generation and comprehension tasks. These findings confirm that “strategy over scale” is a viable approach for AI-assisted hardware design: properly architected agentic frameworks allow smaller models to succeed on tasks traditionally reserved for large models. Beyond immediate performance gains, this opens the door to sustainable, task-specific SLMs that maintain high accuracy and reduce the AI energy footprint.

\printbibliography

@article{patterson2022carbon,
  title={Carbon emissions and large neural network training},
  author={Patterson, David and Gonzalez, Joseph and Le, Quoc and Liang, Chen and Munguia, Lluis-Miquel and Rothchild, Daniel and So, David and Texier, Maud and Dean, Jeff},
  journal={arXiv preprint arXiv:2104.10350},
  year={2022}
}

@inproceedings{liu2023verilogeval,
  title={VerilogEval: Evaluating large language models for Verilog code generation},
  author={Liu, Mingjie and Pinckney, Nathaniel and Khailany, Brucek and Ren, Haoxing},
  booktitle={2023 IEEE/ACM International Conference on Computer Aided Design (ICCAD)},
  pages={1--8},
  year={2023},
  organization={IEEE}
}

@article{thakur2023verigen,
  title={Verigen: A large language model for Verilog code generation},
  author={Thakur, Shailja and Ahmad, Baleegh and Pearce, Hammond and Tan, Benjamin and Dolan-Gavitt, Brendan and Karri, Ramesh and Garg, Siddharth},
  journal={arXiv preprint arXiv:2308.00708},
  year={2023}
}

@article{spec2rtl2025,
  title={Spec2RTL-Agent: Automated hardware code generation from complex specifications using LLM agent systems},
  author={Yu, Zhongzhi and Liu, Mingjie and Zimmer, Michael and Lin, Yingyan and Liu, Yong and Ren, Haoxing},
  journal={arXiv preprint arXiv:2506.13905},
  year={2025}
}

@article{uvllm2024,
  title={UVLLM: An automated universal RTL verification framework using LLMs},
  author={Hu, Yuchen and Ye, Junhao and Xu, Ke and Sun, Jialin and Zhang, Shiyue and Jiao, Xinyao and Pan, Dingrong and Zhou, Jie and Wang, Ning and Shan, Weiwei and others},
  journal={arXiv preprint arXiv:2411.16238},
  year={2024}
}

@misc{cvdp2024,
  title={Comprehensive Verilog design problems benchmark},
  author={{NVIDIA Corporation}},
  year={2024},
  howpublished={\url{https://github.com/NVlabs/verilog-eval}}
}

@article{touvron2023llama,
  title={LLaMA: Open and efficient foundation language models},
  author={Touvron, Hugo and Lavril, Thibaut and Izacard, Gautier and Martinet, Xavier and Lachaux, Marie-Anne and Lacroix, Timoth{\'e}e and Rozi{\`e}re, Baptiste and Goyal, Naman and Hambro, Eric and Azhar, Faisal and others},
  journal={arXiv preprint arXiv:2302.13971},
  year={2023}
}

@article{betterv2024,
  title={BetterV: Controlled Verilog generation with discriminative guidance},
  author={Tsai, Yun-Da and Liu, Mingjie and Ren, Haoxing},
  journal={arXiv preprint arXiv:2402.03255},
  year={2024}
}

@article{chipnemo2023,
  title={ChipNeMo: Domain-adapted LLMs for chip design},
  author={Liu, Mingjie and Ene, Teodor-Dumitru and Kirby, Robert and Cheng, Chris and Pinckney, Nathaniel and Liang, Rongjian and Alben, Jonah and Anand, Himyanshu and Banerjee, Sanmitra and Bayat, Ismet and others},
  journal={arXiv preprint arXiv:2311.00176},
  year={2023}
}

@article{rtlcoder2024,
  title={RTLCoder: Fully open-source and efficient LLM-assisted RTL code generation technique},
  author={Liu, Shang and Fang, Wenji and Lu, Yao and Wang, Jing and Zhang, Qijun and Zhang, Hongce and Xie, Zhiyao},
  journal={IEEE Transactions on Computer-Aided Design of Integrated Circuits and Systems},
  year={2024}
}

@article{ho2024verilogcoder,
  title={VerilogCoder: Autonomous Verilog coding agents with graph-based planning and abstract syntax tree (AST)-based waveform tracing tool},
  author={Ho, Chia-Tung and Ren, Haoxing and Khailany, Brucek},
  journal={arXiv preprint arXiv:2408.08927},
  year={2024}
}

@article{fu2023gpt4aigchip,
  title={GPT4AIGChip: Towards next-generation AI accelerator design automation via large language models},
  author={Fu, Yun and Zhang, Yibo and Yu, Zhongzhi and Li, Shaowen and Ye, Zhenya and Li, Cheng and Wan, Chenhao and Lin, Yonggan},
  journal={arXiv preprint arXiv:2309.10730},
  year={2023}
}

@inproceedings{wang2024chatcpu,
  title={ChatCPU: An agile CPU design and verification platform with LLM},
  author={Wang, Xiao and Wan, Guo-Wei and Wong, Sin-Zhong and Zhang, Li and Liu, Tao and Tian, Qiang and Ye, Jiaxin},
  booktitle={Proceedings of the 61st ACM/IEEE Design Automation Conference},
  pages={1--6},
  year={2024}
}

@misc{phi3,
  title={Phi-3 technical report: A highly capable language model locally on your phone},
  author={{Microsoft Research}},
  year={2024},
  howpublished={\url{https://azure.microsoft.com/en-us/products/ai-services/phi-3}}
}

@misc{deepseek2024,
  title={DeepSeek-R1: Advancing code intelligence with reinforcement learning},
  author={{DeepSeek AI}},
  year={2024},
  howpublished={\url{https://github.com/deepseek-ai/DeepSeek-Coder}}
}

@misc{granite4_2025,
  title={IBM Granite 4.0: hyper-efficient, high performance hybrid models for enterprise},
  author={{IBM}},
  year={2025},
  howpublished={\url{https://www.ibm.com/new/announcements/ibm-granite-4-0-hyper-efficient-high-performance-hybrid-models}}
}

@misc{qwen2024,
  title={Qwen-Coder: Specialized coding models},
  author={{Alibaba Cloud}},
  year={2024},
  howpublished={\url{https://github.com/QwenLM/Qwen}}
}

@misc{smollm2024,
  title={SmolLM2: Compact language models for edge deployment},
  author={{Hugging Face}},
  year={2024},
  howpublished={\url{https://huggingface.co/HuggingFaceTB/SmolLM2-1.7B}}
}

@article{kaplan2020scaling,
  title={Scaling laws for neural language models},
  author={Kaplan, Jared and McCandlish, Sam and Henighan, Tom and Brown, Tom B and Chess, Benjamin and Child, Rewon and Gray, Scott and Radford, Alec and Wu, Jeffrey and Amodei, Dario},
  journal={arXiv preprint arXiv:2001.08361},
  year={2020}
}

@article{autogen2023,
  title={AutoGen: Enabling next-gen LLM applications via multi-agent conversation},
  author={Wu, Qingyun and Bansal, Gagan and Zhang, Jieyu and Wu, Yiran and Li, Beibin and Zhu, Erkang and Jiang, Li and Zhang, Xiaoyun and Zhang, Shaokun and Liu, Jiale and others},
  journal={arXiv preprint arXiv:2308.08155},
  year={2023}
}

@article{agentcoder2023,
  title={AgentCoder: Multi-agent-based code generation with iterative testing and optimisation},
  author={Huang, Dong and Zhang, Jie Ming and Luck, Michael and Bu, Qingwen and Qing, Yichao and Cui, Heming},
  journal={arXiv preprint arXiv:2312.13010},
  year={2023}
}

@article{autoagents2023,
  title={AutoAgents: A framework for automatic agent generation},
  author={Chen, Guangyao and Dong, Siwei and Shu, Yu and Zhang, Ge and Sesay, Jaward and Karlsson, B{\"o}rje F and Fu, Jie and Shi, Yemin},
  journal={arXiv preprint arXiv:2309.17288},
  year={2023}
}

@article{autosafecoder2024,
  title={AutoSafeCoder: A multi-agent framework for securing LLM code generation through static analysis and fuzz testing},
  author={Nunez, Ana and Islam, Nafis Tanveer and Jha, Sumit Kumar and Najafirad, Peyman},
  journal={arXiv preprint arXiv:2409.10737},
  year={2024}
}

@article{semiconductor_energy2023,
  title={The growing energy footprint of the semiconductor industry},
  author={{Semiconductor Industry Association}},
  journal={SIA Report},
  year={2023}
}

@article{saadfalcon2025intelligence,
  title={Intelligence per Watt: Measuring Intelligence Efficiency of Local AI},
  author={Saad-Falcon, Jon and Narayan, Avanika and Akengin, Hakki Orhun and Griffin, J. Wes and Shandilya, Herumb and Gamarra Lafuente, Adrian and Goel, Medhya and Joseph, Rebecca and Natarajan, Shlok and Guha, Etash Kumar and Zhu, Shang and Athiwaratkun, Ben and Hennessy, John and Mirhoseini, Azalia and R{\'e}, Christopher},
  journal={arXiv preprint arXiv:2511.07885},
  year={2025}
}

@article{lester2021power,
  title={The power of scale for parameter-efficient prompt tuning},
  author={Lester, Brian and Al-Rfou, Rami and Constant, Noah},
  journal={arXiv preprint arXiv:2104.08691},
  year={2021}
}

@article{li2021prefix,
  title={Prefix-tuning: Optimizing continuous prompts for generation},
  author={Li, Xiang Lisa and Liang, Percy},
  journal={arXiv preprint arXiv:2101.00190},
  year={2021}
}

@misc{belcak2025smalllanguagemodelsfuture,
      title={Small Language Models are the Future of Agentic AI}, 
      author={Peter Belcak and Greg Heinrich and Shizhe Diao and Yonggan Fu and Xin Dong and Saurav Muralidharan and Yingyan Celine Lin and Pavlo Molchanov},
      year={2025},
      eprint={2506.02153},
      archivePrefix={arXiv},
      primaryClass={cs.AI},
      url={https://arxiv.org/abs/2506.02153}, 
}

@misc{jegham2025hungryaibenchmarkingenergy,
      title={How {H}ungry is {AI}? {B}enchmarking Energy, Water, and Carbon Footprint of LLM Inference}, 
      author={Nidhal Jegham and Marwan Abdelatti and Chan Young Koh and Lassad Elmoubarki and Abdeltawab Hendawi},
      year={2025},
      eprint={2505.09598},
      archivePrefix={arXiv},
      primaryClass={cs.CY},
      url={https://arxiv.org/abs/2505.09598}, 
}

@article{xu2025large,
  title={Large Language Models (LLMs) for Electronic Design Automation (EDA)},
  author={Xu, Kangwei and Schwachhofer, Denis and Blocklove, Jason and Polian, Ilia and Domanski, Peter and Pfl{\"u}ger, Dirk and Garg, Siddharth and Karri, Ramesh and Sinanoglu, Ozgur and Knechtel, Johann and others},
  journal={arXiv preprint arXiv:2508.20030},
  year={2025}
}

@inproceedings{firouzi2025chipmnd,
  title={ChipMnd: LLMs for Agile Chip Design},
  author={Firouzi, Farshad and Pan, David Z and Gu, Jiaqi and Farahani, Bahar and Chaudhuri, Jayeeta and Yin, Ziang and Ma, Pingchuan and Domanski, Peter and Chakrabarty, Krishnendu},
  booktitle={2025 IEEE 43rd VLSI Test Symposium (VTS)},
  pages={1--10},
  year={2025},
  organization={IEEE}
}

@article{zhao2025codev,
  title={Codev: Empowering llms with hdl generation through multi-level summarization},
  author={Zhao, Yang and Huang, Di and Li, Chongxiao and Jin, Pengwei and Song, Muxin and Xu, Yinan and Nan, Ziyuan and Gao, Mingju and Ma, Tianyun and Qi, Lei and others},
  journal={IEEE Transactions on Computer-Aided Design of Integrated Circuits and Systems},
  year={2025},
  publisher={IEEE}
}

@article{pan2025survey,
  title={A survey of research in large language models for electronic design automation},
  author={Pan, Jingyu and Zhou, Guanglei and Chang, Chen-Chia and Jacobson, Isaac and Hu, Jiang and Chen, Yiran},
  journal={ACM Transactions on Design Automation of Electronic Systems},
  volume={30},
  number={3},
  pages={1--21},
  year={2025},
  publisher={ACM New York, NY}
}

@article{wang2025mcp4eda,
  title={MCP4EDA: LLM-Powered Model Context Protocol RTL-to-GDSII Automation with Backend Aware Synthesis Optimization},
  author={Wang, Yiting and Ye, Wanghao and He, Yexiao and Chen, Yiran and Qu, Gang and Li, Ang},
  journal={arXiv preprint arXiv:2507.19570},
  year={2025}
}

@misc{ghose2025orfsagenttoolusingagentschip,
      title={ORFS-agent: Tool-Using Agents for Chip Design Optimization}, 
      author={Amur Ghose and Andrew B. Kahng and Sayak Kundu and Zhiang Wang},
      year={2025},
      eprint={2506.08332},
      archivePrefix={arXiv},
      primaryClass={cs.AI},
      url={https://arxiv.org/abs/2506.08332}, 
}

@INPROCEEDINGS{orfs_agent,
  author={Wu, Bing-Yue and Sharma, Utsav and Rovinski, Austin and Chhabria, Vidya A.},
  booktitle={2025 IEEE International Conference on LLM-Aided Design (ICLAD)}, 
  title={OpenROAD Agent: An Intelligent Self-Correcting Script Generator for OpenROAD}, 
  year={2025},
  volume={},
  number={},
  pages={16-22},
  doi={10.1109/ICLAD65226.2025.00039}}

@article{xu2025revolution,
  title={Revolution or Hype? Seeking the Limits of Large Models in Hardware Design},
  author={Xu, Qiang and Stok, Leon and Drechsler, Rolf and Wang, Xi and Zhang, Grace Li and Markov, Igor L},
  journal={arXiv preprint arXiv:2509.04905},
  year={2025}
}

@misc{deng2025chipaligninstructionalignmentlarge,
      title={ChipAlign: Instruction Alignment in Large Language Models for Chip Design via Geodesic Interpolation}, 
      author={Chenhui Deng and Yunsheng Bai and Haoxing Ren},
      year={2025},
      eprint={2412.19819},
      archivePrefix={arXiv},
      primaryClass={cs.AR},
      url={https://arxiv.org/abs/2412.19819}, 
}

@inproceedings{li2025specllm,
  title={Specllm: Exploring generation and review of vlsi design specification with large language model},
  author={Li, Mengming and Fang, Wenji and Zhang, Qijun and Xie, Zhiyao},
  booktitle={2025 International Symposium of Electronics Design Automation (ISEDA)},
  pages={749--755},
  year={2025},
  organization={IEEE}
}

@article{meng2023unlocking,
  title={Unlocking hardware security assurance: The potential of LLMs},
  author={Meng, Xingyu and Srivastava, Amisha and Arunachalam, Ayush and Ray, Avik and Silva, Pedro Henrique and Psiakis, Rafail and Makris, Yiorgos and Basu, Kanad},
  journal={arXiv preprint arXiv:2308.11042},
  year={2023}
}

@String{Computer = "{IEEE} Computer" }

@misc{pulavarthi2025assertionbenchbenchmarkevaluatelargelanguage,
      title={AssertionBench: A Benchmark to Evaluate Large-Language Models for Assertion Generation}, 
      author={Vaishnavi Pulavarthi and Deeksha Nandal and Soham Dan and Debjit Pal},
      year={2025},
      eprint={2406.18627},
      archivePrefix={arXiv},
      primaryClass={cs.SE},
      url={https://arxiv.org/abs/2406.18627}, 
}

\end{document}